\documentclass[final,onefignum,onetabnum]{siamart220329}

\usepackage{braket,amsfonts}
\usepackage{array}
\usepackage{tabularray}
\usepackage{cleveref}
\usepackage[caption=false]{subfig}
\usepackage{pgfplots}

\newsiamthm{claim}{Claim}
\newsiamremark{remark}{Remark}
\newsiamremark{hypothesis}{Hypothesis}
\crefname{hypothesis}{Hypothesis}{Hypotheses}

\usepackage{algorithmic}

\usepackage{graphicx,epstopdf}

\Crefname{ALC@unique}{Line}{Lines}

\usepackage{amsopn}

\usepackage{xspace}
\usepackage{bold-extra}
\usepackage[most]{tcolorbox}

\colorlet{texcscolor}{blue!50!black}
\colorlet{texemcolor}{red!70!black}
\colorlet{texpreamble}{red!70!black}
\colorlet{codebackground}{black!25!white!25}

\lstdefinestyle{siamlatex}{%
  style=tcblatex,
  texcsstyle=*\color{texcscolor},
  texcsstyle=[2]\color{texemcolor},
  keywordstyle=[2]\color{texemcolor},
  moretexcs={cref,Cref,maketitle,mathcal,text,headers,email,url},
}

\tcbset{%
  colframe=black!75!white!75,
  coltitle=white,
  colback=codebackground, 
  colbacklower=white, 
  fonttitle=\bfseries,
  arc=0pt,outer arc=0pt,
  top=1pt,bottom=1pt,left=1mm,right=1mm,middle=1mm,boxsep=1mm,
  leftrule=0.3mm,rightrule=0.3mm,toprule=0.3mm,bottomrule=0.3mm,
  listing options={style=siamlatex}
}

\newtcblisting[use counter=example]{example}[2][]{%
  title={Example~\thetcbcounter: #2},#1}

\newtcbinputlisting[use counter=example]{\examplefile}[3][]{%
  title={Example~\thetcbcounter: #2},listing file={#3},#1}

\DeclareTotalTCBox{\code}{ v O{} }
{ 
  fontupper=\ttfamily\color{black},
  nobeforeafter,
  tcbox raise base,
  colback=codebackground,colframe=white,
  top=0pt,bottom=0pt,left=0mm,right=0mm,
  leftrule=0pt,rightrule=0pt,toprule=0mm,bottomrule=0mm,
  boxsep=0.5mm,
  #2}{#1}

\patchcmd\newpage{\vfil}{}{}{}
\usepackage{footmisc} 

\begin{tcbverbatimwrite}{tmp_\jobname_header.tex}
\title{Estimating the Distribution of Parameters in Differential Equations with Repeated Cross-Sectional Data\thanks{Submitted to the editors \today.
\funding{Hyung Ju Hwang was supported by the National Research Foundation of Korea(NRF) grant funded by the Korea government(MSIT) (No. RS-2023-00219980 and RS-2022-00165268) and by Institute for Information \& Communications Technology Promotion (IITP) grant funded by the Korea government(MSIP) (No.2019-0-01906, Artificial Intelligence Graduate School Program (POSTECH). Hyeontae Jo was supported by a Korea University Grant.
}}}

\author{Hyeontae Jo\thanks{First author. Department of Mathematics, Korea University Sejong Campus, Sejong 30019, Republic of Korea and Biomedical Mathematics Group, Pioneer Research Center for Mathematical and Computational Sciences, Institute for Basic Science, Daejeon, 34126, Republic of Korea(\email{korea\_htj@korea.ac.kr}).}
\and Sung Woong Cho\thanks{Equal contribution. Stochastic Analysis and Application Research Center, Korea Advanced Institute of Science and Technology, Daejeon 34141, Republic of Korea (\email{swcho95kr@kaist.ac.kr}).}
\and Hyung Ju Hwang\thanks{Corresponding Author. Department of Mathematics \& Graduate School of AI, Pohang University of Science and Technology, Pohang 37673, Republic of Korea (\email{hjhwang@postech.ac.kr}).}}

\end{tcbverbatimwrite}
\input{tmp_\jobname_header.tex}

\ifpdf
\hypersetup{ pdftitle={Guide to Using  SIAM'S \LaTeX\ Style} }
\fi
\begin{document}
\maketitle

\begin{tcbverbatimwrite}{tmp_\jobname_abstract.tex}
\begin{abstract}
Differential equations are pivotal in modeling and understanding the dynamics of various systems, offering insights into their future states through parameter estimation fitted to time series data. In fields such as economy, politics, and biology, the observation data points in the time series are often independently obtained (i.e., Repeated Cross-Sectional (RCS) data). With RCS data, we found that traditional methods for parameter estimation in differential equations, such as using mean values of time trajectories or Gaussian Process-based trajectory generation, have limitations in estimating the shape of parameter distributions, often leading to a significant loss of data information. To address this issue, we introduce a novel method, Estimation of Parameter Distribution (EPD), providing accurate distribution of parameters without loss of data information. EPD operates in three main steps: generating synthetic time trajectories by randomly selecting observed values at each time point, estimating parameters of a differential equation that minimize the discrepancy between these trajectories and the true solution of the equation, and selecting the parameters depending on the scale of discrepancy. We then evaluated the performance of EPD across several models, including exponential growth, logistic population models, and target cell-limited models with delayed virus production, demonstrating its superiority in capturing the shape of parameter distributions. Furthermore, we applied EPD to real-world datasets, capturing various shapes of parameter distributions rather than a normal distribution. These results effectively address the heterogeneity within systems, marking a substantial progression in accurately modeling systems using RCS data. Thus, EPD marks a significant advancement in accurately modeling systems with RCS data, enabling a deeper understanding of system dynamics and parameter variability.
\end{abstract}

\begin{keywords}
differential equation, parameter estimation, repeated cross-sectional data, distribution of parameters
\end{keywords}

\begin{MSCcodes}
62G07, 62G09, 62P10, 65D10
\end{MSCcodes}
\end{tcbverbatimwrite}
\input{tmp_\jobname_abstract.tex}

\section{Introduction}
\label{sec:intro}

Differential equations play a crucial role in modeling the evolution of various systems, offering scientific and mechanistic insights into physical and biological phenomena and enabling predictions of their future states. These phenomena can be analyzed by parameters of the differential equation that fit its solutions to time series data. However, in systems such as economy, politics, or biology, data observations are often Repeated Cross-Sectional (RCS) (i.e., data is collected over time measuring the same variables with different samples or populations at each time point) \cite{beck2007random,beck2011modeling,pan2022repeated, stevens2018hbsc, bryman2016social}. For example, Sara, et al. analyzed the degree of tumor size suppression over time in rats with different types of drugs, using an exponential growth model \cite{lundsten2020radiosensitizer}. As mice died during the experiment, observation data from the experiment can not be connected per time (i.e., RCS data). For other cases, Jeong et al. utilized time series data on the PER protein levels in Drosophila to analyze neuron-dependent molecular properties \cite{jeong2022systematic}. However, measuring PER levels at each time point necessitated the sacrifice of the flies, thus limitations in the collection of RCS data inevitably happened. RCS data also includes regular surveys in society that collect the changing opinions of different individuals. Public polls by Gallup, the Michigan Survey of Consumers \cite{clarke2005men, hopkins2012whose}, records of congressional roll calls \cite{lebo2007strategic}, Supreme Court cases \cite{segal2002supreme}, and presidential public remarks \cite{wood2009myth} are all examples of RCS data.

Fitting the parameters with cross-sectional data or time-series data is feasible with classical optimization methods, yet handling RCS data poses a significant challenge. While several methods have been used, their applicability is constrained. For example, one common method involves using the mean values at each time point for parameter estimation \cite{jeong2022systematic}. While this simplifies the analysis of RCS data, it significantly reduces the data information. To preserve the data information, Gaussian Process-based time series generation (GP) is utilized for model calibration. Specifically, GP produces continuous-time trajectories through the mean and covariance of RCS data, enabling us to use traditional parameter estimation techniques. Nonetheless, since the GP method relies solely on the mean and covariance, the estimation results from GP-based algorithms tend to be unimodal \cite{smith2018host, chung2019parameter, zhang2020improved, binois2021hetgp}. Thus, this approach can fail when the underlying distribution is not unimodal, potentially leading to an incorrect estimation of the shape of parameter distributions and a loss of data information \cite{jo2024density}.

\begin{figure}[tbhp]
  \centering
  \includegraphics[width=1\textwidth]{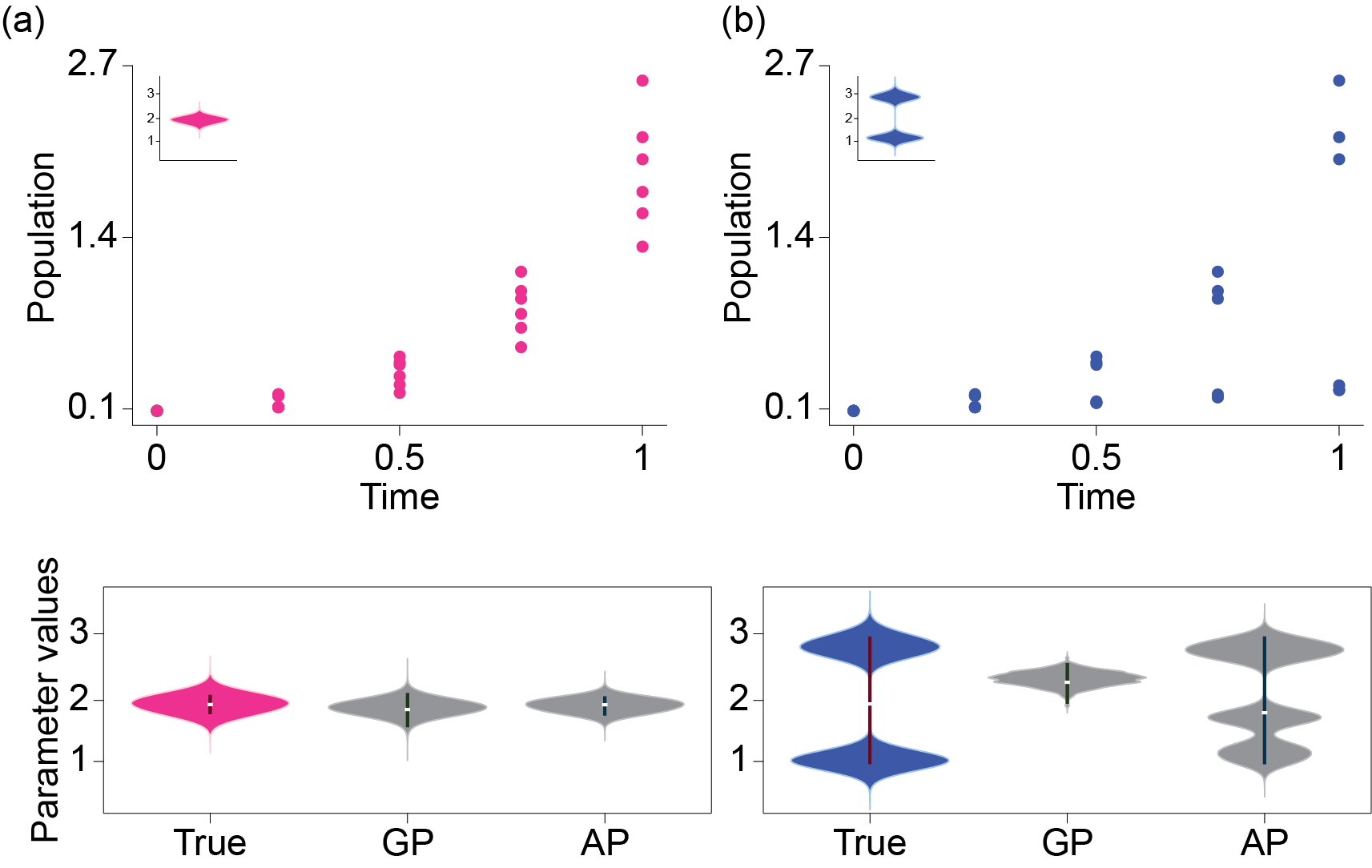}
  \caption{{\bf Parameter estimation in the exponential growth model with Repeated Cross-Sectional (RCS) data.} An exponential growth model $y’(t)=ay(t)$ represents the amount of population, y(t), changes over time, $t$. We then estimated parameter $a$ that can fit the model to a given RCS data (a-b). When the true parameter distribution of $a$ is unimodal (a, top-panel), corresponding RCS data is generated by parameters $a$, and populations per time do not diverge (a, top).In this case, previous methods, such as Gaussian Process (GP) or All Possible combinations (AP), can estimate true parameter distributions (bottom) (a, bottom). When the true parameter distribution of $a$ is bimodal (b, top-penal), populations per time diverge (b, top). In this case, previous methods fail to estimate the shape of true parameters (b, bottom).}
  \label{fig:1}
\end{figure}

In this paper, we introduce a novel approach, Estimation of Parameter Distribution (EPD), to infer parameter distributions from RCS data in systems modeling. Our proposed method stands out for its ability to accurately and precisely determine the parameter distributions in various systems through the following two steps: In the first step, we randomly choose one observed value for every time point, creating an artificial time trajectory. Next, we estimate the parameters $p$ of the differential equation that minimize the difference between the time trajectory and its solution, denoted by $L(p)$. In the second step, by repeating the first step $N$ times, we obtain a collection of parameter sets $p_n$ along with their respective differences $L(p_n)$, for $n=1,...,N$. Next, we define the probability that each $p$ came from the true parameter distribution based on the $\{L(p_n)\}$, and draw the distribution by collecting only $p$ selected based on their probability values. Through this process, we show that EPD accurately captures true parameter distributions for the following models: 1) exponential growth, 2) logistic population models \cite{yada2023few}, and 3) target cell-limited model with delayed virus production \cite{chung2019parameter,smith2018influenza,myers2021dynamically},

In this study, we found that previous methods fail to estimate the distribution of parameters when the distributions do not follow a normal distribution, leading to the loss of data information \cref{fig:1}. To address this, we developed an EPD that can accurately estimate the shape of the parameter distribution, resulting in a more comprehensive, deeper, and better understanding of the data. Hence, by analyzing the shape of these parameter distributions, we can deduce the underlying circumstances and dynamics of the system in question.

\section{The parameter estimation problem: a general description of our problems and suggested methods}
\label{sec2:Parameter estimation problem}
\subsection{Problem formulation}
\label{subsec2.1:problem formulation} 
\begin{figure}[tbhp]
  \centering
  \includegraphics[width=1\textwidth]{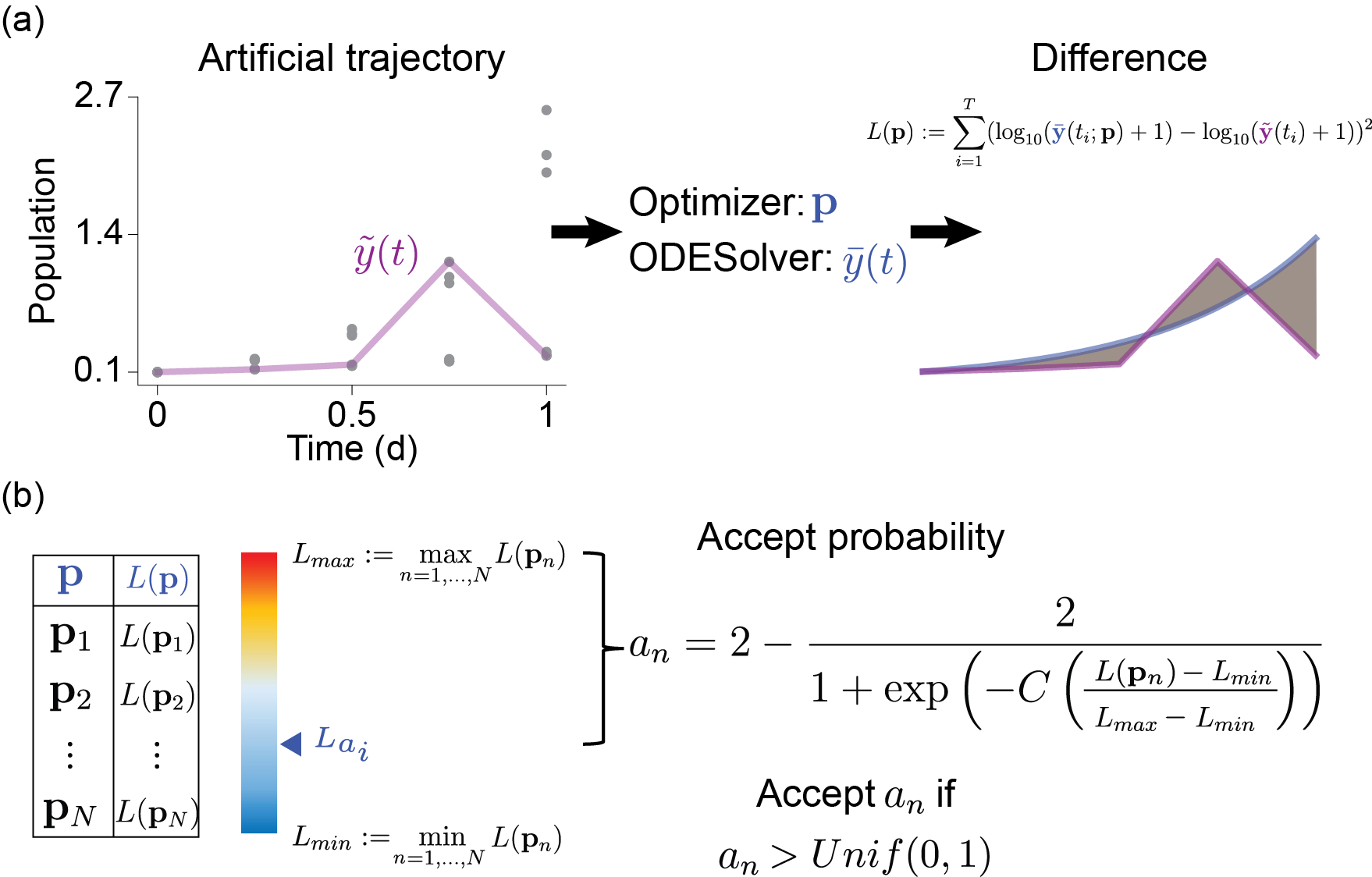}
  \caption{{\bf Parameter estimation in the exponential growth model with Repeated Cross-Sectional (RCS) data.} An exponential growth model $y’(t)=ay(t)$ represents the amount of population, y(t), changing over time, $t$. We then estimated parameter $a$ that can fit the model to a given RCS data (a-b). When the true parameter distribution of $a$ is unimodal (a, top-panel), corresponding RCS data is generated by parameters $a$, and populations per time do not diverge (a, top). In this case, previous methods, such as Gaussian Process (GP) or All Possible combinations (AP), can estimate true parameter distributions (bottom) (a, bottom). When the true parameter distribution of $a$ is bimodal (b, top-penal), populations per time diverge (b, top). In this case, previous methods fail to estimate the shape of true parameters (b, bottom).}
  \label{fig:2}
\end{figure}

We propose the method for estimating the distribution of parameters within the time evolutionary differential equation (ODE), represented as:
\begin{equation}\label{equation:ODE}
\mathbf{y}'(t) = f[\mathbf{y}(t), \mathbf{p}, t]
\end{equation}
where $\mathbf{y} = \mathbf{y}(t) \in \mathbb{R}^{n_y}$ represents the nonnegative population size with dimension $n_y$ at time $t$. The parameter set $\mathbf{p}\in\mathbb{R}^{n_p}$ represents biological or physical properties such as the growth rate or the carrying capacity for the population, respectively. The problem is to estimate the distribution of the parameter $p$ when the observation data corresponding to $y(t)$, $Y$, is given as RCS data (\cref{fig:2}a, left). Specifically, $Y$ includes a set of observed data points at each time step $t_i$, $Y_i$, for $i=1, 2, \cdots, T$, where $T$ is the total count of time steps. Each $Y_i$ includes $J$ observed data points at time $t_i$, (i.e., $Y_i=\{y_1(t_i), y_2(t_i),..., y_J(t_i)\}$). As the data $Y$ not only have different observation values per time $t_i$ but also are independent (i.e., RCS data), each $y_i(t_j)$ can have different parameter values $\mathbf{p}$. That is, we try to estimate the distribution of parameter $\mathbf{p}$, rather than a single fixed value $\mathbf{p}$.

\subsection{Development of EPD, estimating the distribution of parameters}
\label{subsec2.2:Development of EPD}
 To estimate parameters corresponding to  RCS data, we construct $N$ artificial trajectories, denoted as $\{\tilde{y}_n\}_{n=1}^{N}$, aligned with specific time points $\{t_i\}_{i=1}^{T}$ (\cref{fig:2}a, right). Specifically, for each $t_i$, we randomly chose one observation value, $y_{i_j}$, from $Y_i$ with $1\leq i_j \leq J$. We assume this selection probability to be equivalent. We then consider $\{y_{i_j}\}$ as the artificial time trajectory $\tilde{y}$. Repeating this $N$ times, we can obtain $n$ artificial time trajectories $\{\tilde{y}_n\}_{n=1}^{N}$. Remarkably, the likelihood of choosing the trajectory $\tilde{y}_n(\cdot)$, given the observations $\{Y_i\}_{i=1}^{T}$, can be formulated as follows:
\begin{align*}
P(\{\tilde{y}_{n}|\{Y_i\}_{i=1}^{T}\}) & = P(\tilde{y}_{n}(t_1)=y_{1_j}, \tilde{y}_{n}(t_2)=y_{2_j}, \ldots, 
\tilde{y}_{n}(t_n)=y_{n_j}) \\
& = \Pi_{i=1}^{T}P(\tilde{y}_{n}(t_i)=y_{i_j}))=\Pi_{i=1}^{T} 1/J = (1/J)^T.
\end{align*}

Next, we estimate the parameter $\mathbf{p}_n$ corresponding to $n$-th artificial trajectory $\tilde{y}_n$. For this, our method utilizes a deterministic least squares optimization to reduce the difference between $\tilde{y}_n$ and solution of \eqref{equation:ODE} with $\mathbf{p}_n$, $\bar{y}_n$. Specifically, we solve equation \eqref{equation:ODE} with $\mathbf{p_n}$ to obtain the corresponding trajectory $\bar{y}_n$ through LSODA algorithm \cite{hindmarsh1983odepack, virtanen2020scipy}, as follows:
\begin{equation}\label{eq:reconstruct_trajectory}
\bar{\mathbf{y}}_n(t; \mathbf{p})={\mathbf{y}}(f, \mathbf{y}(0), t;\mathbf{p}), \forall \mathbf{p}.
\end{equation}
The objective function for the optimization is defined as the sum of squared deviations between the logarithmically transformed observed data and the model predictions. For a given trajectory $\tilde{\mathbf{y}}_{n}$, it can be formulated as:

\begin{equation}\label{eq:objective}
L(\mathbf{p})=\sum_{i=1}^{T}(\log_{10}(\bar{\mathbf{y}}_n (t_i;\mathbf{p})+1)-\log_{10}(\tilde{\mathbf{y}}_n (t_i)+1))^2.
\end{equation}
To minimize $L(\mathbf{p})$, we employ LMFIT \cite{newville2016lmfit} package in Python to apply the least square algorithm. Through the numerical solution of the ODE for each $\tilde{y}_n$, we obtain the set of parameters $\mathbf{p}_n$.

 As $\tilde{y}_n$ are not real continuous observation trajectories but artificial, we need to determine whether each $\tilde{y}_n$ is reasonable or not. For this determination, we create the accept probability $a_n$, which depends on how well the model \eqref{equation:ODE} fits with the estimated parameters $\mathbf{p}_n$  (\cref{fig:2}b). The probability $a_n$ is calculated via a logistic transformation applied to the previously computed residuals $L(\mathbf{p}_n)$ (representing the goodness of fit) as follows:
\begin{equation}\label{eq:accept_probability}
a_n = 2 - \frac{2}{1 + \exp\left(-C\left(\frac{L(\mathbf{p}_n) - \min_{n} L(\mathbf{p}_n)}{\max_{n} L(\mathbf{p}_n) - \min_{n} L(\mathbf{p}_n)}\right)\right)},
\end{equation}
where $L(\mathbf{p}_n)$ denotes the objective function values in $\eqref{eq:objective}$ for each fit, and $C>0$ represents a scaling factor that can be adjusted for improved accuracy. In contrast to MCMC, which iteratively refines parameter estimates to converge on the posterior distribution, our method decides on accepting or rejecting parameter sets based on their computed likelihood after explicitly minimizing a predefined objective function. Specifically, a parameter set $\mathbf{p}_n$ is accepted if it satisfies:
\begin{equation*}
    a_n>u_n\quad  \text{where} \quad u_n\sim Unif(0, 1),
\end{equation*}
where a set $\{u_n\}_{n=1}^{N}$ is independently sampled from an identical uniform distribution over $n$. It ensures a probabilistic assessment of parameter set acceptance based on their respective goodness of fit. Note that when $C$ equals to zero, all estimated parameters will be accepted. This case will be referred to as All Possible combinations (AP) because it considers every estimated result without further refinement.

\section{Evaluating EPD performance in estimating parameter distributions using simulation datasets}
\label{sec3:Evaluation}
In this section, we evaluated the performance of EPD in estimating the distribution of parameters with simulation data. For this task, we employed three distinct dynamical systems: an exponential growth model for detecting cell dynamics heterogeneity, a logistic regression model for simulating protein generation in \cite{yada2023few}, and a target cell-limited model for understanding virus infection dynamics \cite{baccam2006kinetics, nowak2000virus}. With these evaluations, we show the adaptability and robust potential of EPD in accurately identifying the true parameter distributions and in forecasting system behaviors even in the presence of noise. 
To generate the distribution of $\mathbf{p}$ synthetically, we first consider $H$ distinct centers $\{\mathbf{p}^{center}_h \}_{h=1}^{H}$ which imply the peaks of parameters across various clusters. The large value of $H$ can pose parameter heterogeneity. We conduct uniform random sampling independently within pre-established bounds to generate parameter distribution around these centers. Specifically, we randomly select $S$ values for the parameters $\{\mathbf{p}_{(h-1)S+i}\}_{i=1}^{S}$ around $\mathbf{p}^{center}_{h}$ within their respective bounds as follows:
\begin{equation*}
    \mathbf{p}_{(h-1)s+i}\sim Unif((\mathbf{p}_L)_h, (\mathbf{p}_U)_h),
\end{equation*}
where $Unif((\mathbf{p}_L)_h, (\mathbf{p}_U)_h)$ represents the uniform distribution between $(\mathbf{p}_L)_h$ and $(\mathbf{p}_U)_h$. This results in $HS$ sampled parameter sets which are utilized to construct trajectories and hence generate RCS data related to diverse biological experiment scenarios. 
To generate synthetic RCS data, we resolve the ODE in \eqref{equation:ODE} for each parameter set $\mathbf{p}$. For this, we use the LSODA algorithm, which can adjust the balance between stiff and non-stiff structures of solutions, with initial conditions $\mathbf{y}(0)$ at time $t=0$, 
\begin{equation*}   
    \tilde{\mathbf{y}}(t;\mathbf{p})={\mathbf{y}}(f, \mathbf{y}(0), t;\mathbf{p}), \mathbf{p}\in \{\mathbf{p}_1,\ldots, \mathbf{p}_{HS}\}.
\end{equation*}
We remark that the initial value $\mathbf{y}(0)$ is determined by the experimental setup or an existing dataset. The above ODE solving mechanism yields totally $HS$ trajectories which will be designated as RCS data. Specifically, it is assumed that we can only access $HS$ data points at each observational time point $t_i$ for $i=1, 2, \ldots, T,$ where $T$ is the total number of time steps, instead of a set of fully connected trajectories.

\subsection{EPD can infer the various shapes of underlying parameter distribution of the simple exponential growth model}
\label{subsec3.1:Exponential_growth_model}
\begin{figure}[tbhp]
  \centering
  \includegraphics[width=1\textwidth]{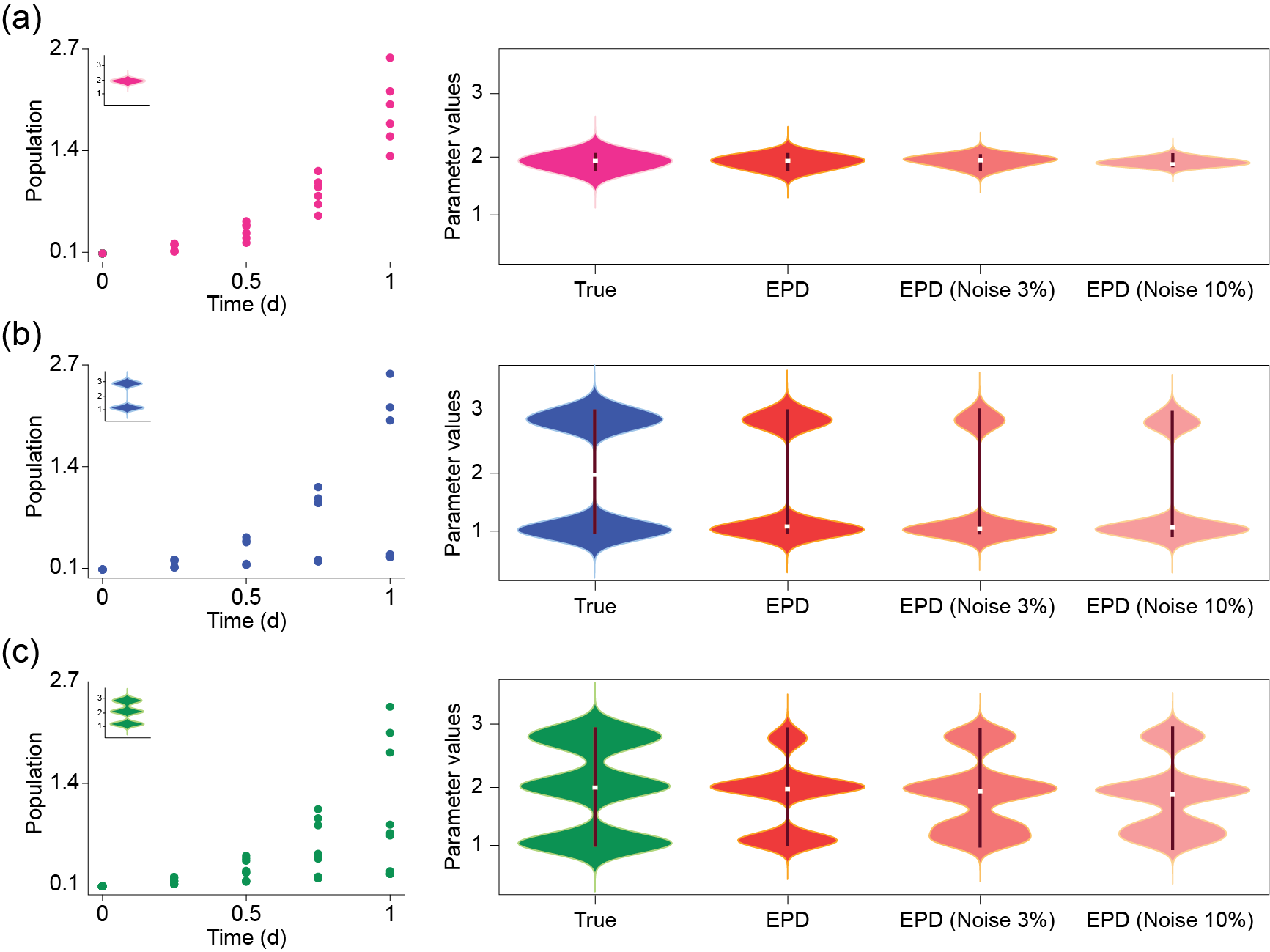}
  \caption{{\bf Accurate estimation of true distributions by EPD in datasets exhibiting unimodal, bimodal, and trimodal parameter distributions within an Exponential Growth Model.} (a) When the true parameter distribution is unimodal, we applied EPD on the observed data (left) and estimated the parameters (right). Notably, EPD remained accurate even when we added 3\% or 10\% multiplicative noise to the data (b, c). Likewise, EPD was confirmed to estimate Bimodal and Trimodal parameter distributions effectively.}
  \label{fig:3}
\end{figure}

The exponential growth model can be used to analyze the growth patterns in population dynamics,  
\begin{equation*}
    y'=ay,
\end{equation*}
where $y(t)$ represents the number of populations at time $t$ and $a$ is the population growth rate. We evaluated the performance of EPD in estimating true parameter distributions that reflect a given dataset through the exponential growth model. For this evaluation, we first generated a simulation dataset through a numerical solver with different five growth rates $a$ obtained from an unimodal distribution (\cref{fig:3}a). Specifically, the simulation dataset consists of five snapshot data at time points $t=0, 0.25, 0.5, 0.75,$ and $1.$ Notably, we assume each observed value is not time-traceable, (i.e., RCS data). To apply this dataset to EPD, we generated 1,000 trajectories with observation values randomly selected at each time point. For each trajectory, we assigned an acceptance probability that reflects the likelihood of the trajectory’s parameters being derived from the true parameter distribution (See Method for details). Through this process, we showed that EPD can accurately estimate the shape of true parameter distribution (i.e., unimodal distribution) (\cref{fig:3}a, right-EPD). Furthermore, EPD also can estimate the same distribution even when the data was subjected to multiplicative noise at levels of 3\% and 10\%, respectively. Subsequently, we extended the evaluation task with different datasets, reflecting different shapes of parameter distributions: a bimodal and a trimodal distribution (\cref{fig:3}(b-c), left), respectively. In each case, EPD consistently inferred the true parameter distributions even when having the noise (\cref{fig:3}(b-c), right). Hence, these simulation results underscore that EPD accurately estimates the true parameter distributions that reflect the dataset. 

\subsection{EPD can infer the various shapes of underlying parameter distribution of the logistic population model}
\label{subsec3.2:Logistic_population_model}
\begin{figure}[tbhp]
  \centering
  \includegraphics[width=1\textwidth]{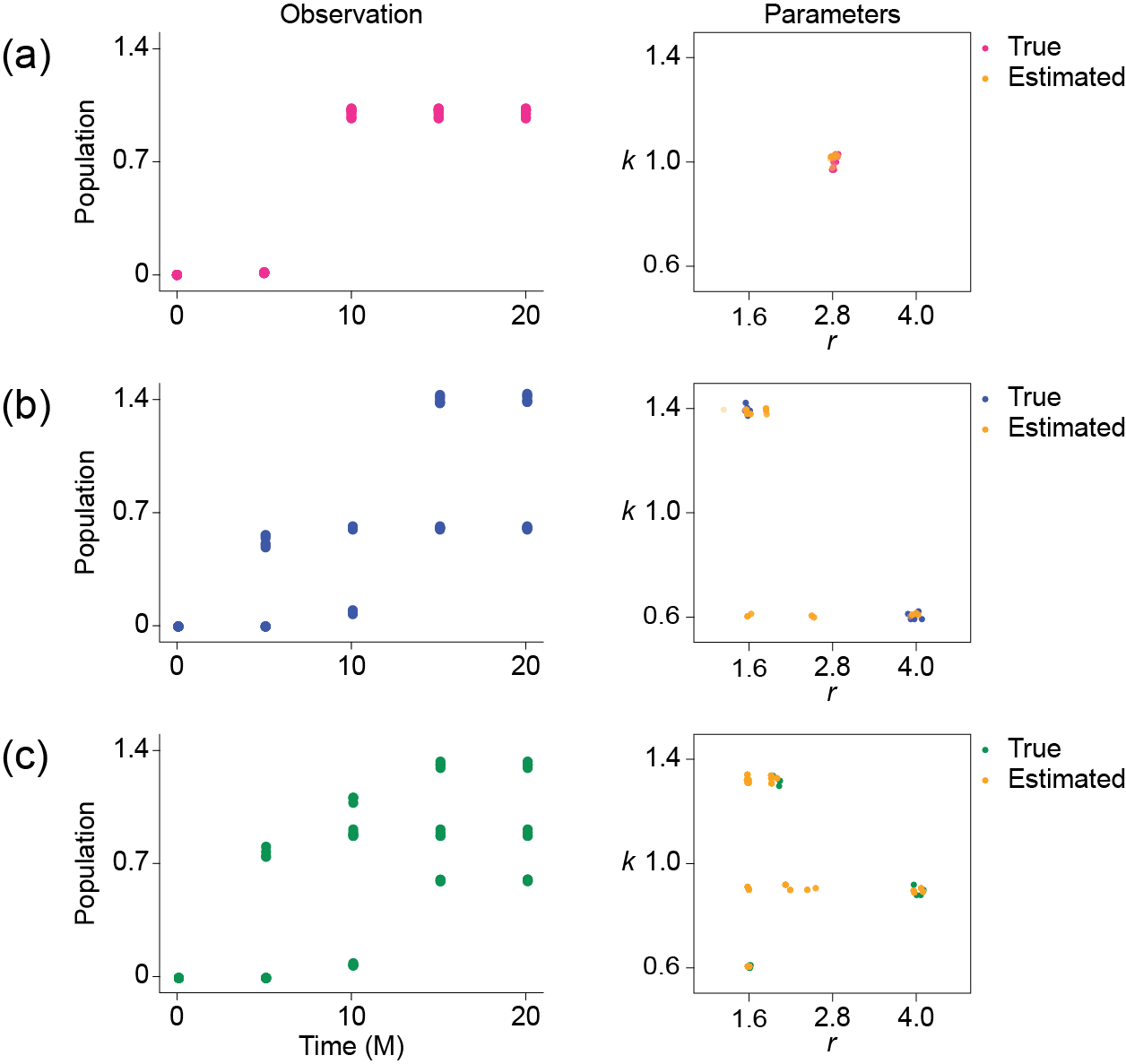}
  \caption{{\bf Estimation of parameter distribution for the logistic growth model.} The model include two parameters $r$ and $k$. For the scenarios of unimodal, bimodal, and trimodal distributions, we used RCS data on the left side and estimated the parameter distribution on the right side.}
  \label{fig:4}
\end{figure}

The logistic population model has been utilized to understand the growth dynamics of the level of protein over time $t$, $y(t)$: 
\begin{equation*}
    y'=ry(1-y/K),
\end{equation*}
where $y$ quantifies protein levels over time, $r$ is the growth rate, and $K$ represents the maximum sustainable population size that the environment can support. To evaluate the estimation performance of EPD with this model, we initially assume an unimodal distribution for the parameters, centered around peaks of $(2.8, 1.0)$ as a true parameter distribution. Subsequently, we generated the 12 numerical solutions for the parameter $(r, K)$ sampled from this distribution. These 12 solutions were then used to record observation data at $t=5, 10, 15,$ and $20$ months, with an initial protein level of \(y(0) = 0.0001\) (\cref{fig:4}a, left). To apply this dataset to EPD, we generated 1,000 trajectories with observation values randomly selected at each time point. Similarly to the results for the first exponential model, EPD demonstrated its efficacy in accurately estimating true parameter distributions (\cref{fig:4}a, right). For the case when the true distribution is bimodal or trimodal, we included sets of parameters near centers ${(4.0, 0.6), (1.6, 1.4)}$ and ${(1.6, 0.6), (4.0, 0.9), (2.0, 1.3)}$, respectively (\cref{fig:4}(b-c), left). After we applied EPD to these datasets separately, we validated EPD can estimate the true parameter distributions regardless of the shape (\cref{fig:4}(b-c), right). Notably, EPD estimated the interpolation of two centers of true parameters. This implies EPD can detect all possible combinations of scenarios for $(r,K)$. That is, in terms of marginal distribution for each parameter, EPD  still shows a high level of accuracy in predicting these distributions.

\subsection{EPD can infer the various shapes of underlying parameter distribution of target cell-limited models with delayed virus production}
\label{subsec3.3:target_cell_limited_model}
\begin{figure}[tbhp]
  \centering
  \includegraphics[width=1\textwidth]{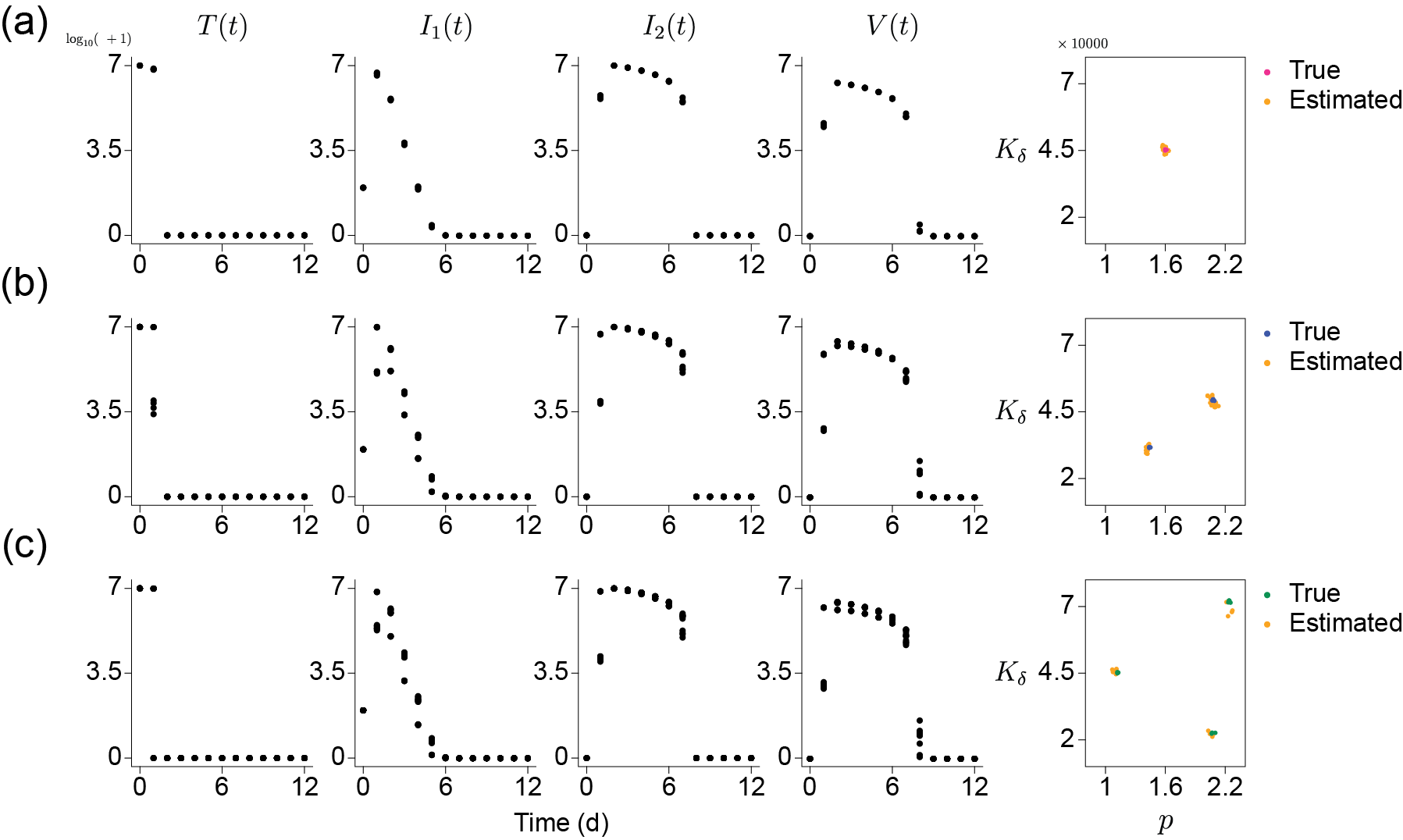}
  \caption{{\bf Estimation results on the parameter distribution for a target cell-limited model with delayed virus production.} This model describes four components $T$, $I_1$, $I_2$, and $V$ over time. We explore three different parameter distributions: unimodal (a), bimodal (b), and trimodal (c). For each case, we used RCS data for four populations (left) and presented estimation results for two parameters $p$ and $K_{\delta}$ (right) among the six parameters contained in the model.}
  \label{fig:5}
\end{figure}

\begin{figure}[t]
  \centering
  \includegraphics[width=\textwidth]{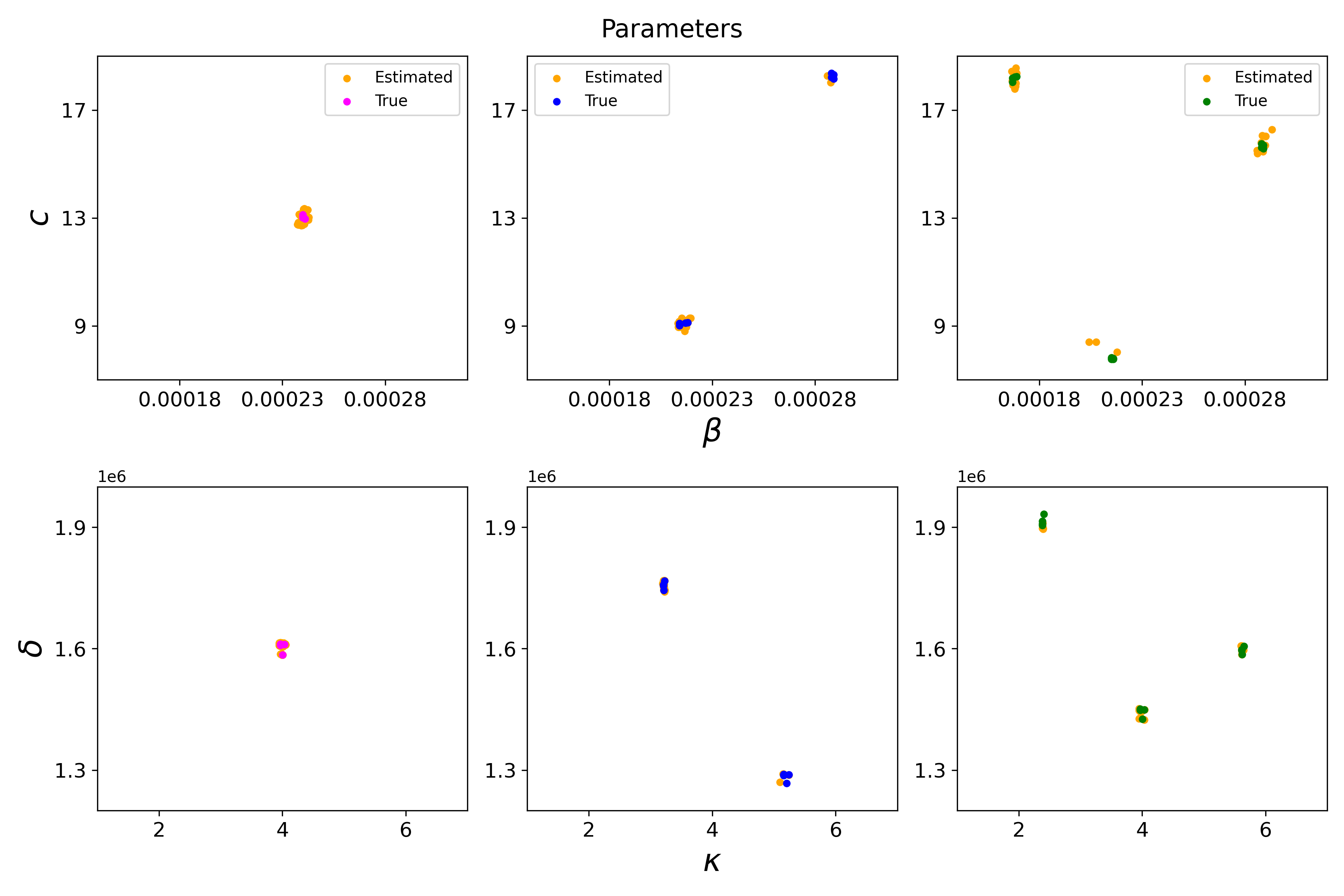}
  \caption{{\bf Estimation results for the parameter distribution within a target cell-limited model based on synthetic data.} We estimated the distributions of $\beta$ $c$, $\kappa$, and $\delta$, corresponding to data in (\cref{fig:5}, left $T(t)-V(t)$), respectively. First, we applied EPD to data generated from parameters that share similar scales (\cref{fig:5}(a), left $T(t)-V(t)$). As a result, EPD is capable of accurately estimating the parameters (Left). Furthermore, With data generated from parameters with different scales (\cref{fig:5}(b-c), left), EPD can infer the true parameter distributions (Middle and Right, respectively). That is, EPD can estimate the true distribution of parameters even when they do not follow the normal distribution. Notably, the prediction does not contain the interpolation of the centers as not previously in the logistic model.}
  \label{fig:6} 
\end{figure}

We additionally performed a benchmark study in estimating the parameter distributions of a target cell-limited model with delayed virus production, characterized by four principal populations: susceptible epithelial cells $T,$ eclipse phase $I_1$, active virus production $I_2$, and the virus population $V.$ With the four variables, the target cell-limited model can be described by the following differential equations:
\begin{align*}
\frac{dT}{dt} &= -\beta T V, \\
\frac{dI_1}{dt} &= \beta T V - \kappa I_1, \\
\frac{dI_2}{dt} &= \kappa I_1-\frac{\delta I2}{K_d+I_2}, \\
\frac{dV}{dt} &= p I_2-cV.
\end{align*}
Specifically, susceptible cells $T$ are infected by the virus proportional to $V$ with proportional constant $\beta$. Subsequently, the infected cells enter the eclipse phase $I_1$ before progressing to active virus production $I_2$. Virus production is regulated at a specific rate $p$ per cell, while the virus $V$ is eliminated at a clearance rate $c$, and infected cells $I_2$ are removed according to the function $I_2/(K_{\delta} + I_2)$, where $K_{\delta}$ represents the half-saturation constant.
To validate the predictive performance of EPD with this model, we obtained a RCS dataset for $T, I_1, I_2,$ and $V$ that is generated by 12 different sets of parameters. Specifically, parameters were chosen near the center ($2.4\times 10^-4$, $1.6$, $13.0$, $4.0$, $1.6\times 10^6$, $4.5\times 10^5$) from \cite{smith2018influenza} (\cref{fig:5}a, left). Using 12 sets of parameters sampled from this distribution, we obtained simulation data over 12 days with initial conditions $[T(0), I_1(0), I_2(0), V(0)] = [10^7, 75, 0, 0].$  We then apply EPD to this dataset, predicting original parameter distributions (\cref{fig:5}a, right). Surprisingly, even when the shape of true parameter distributions is bi- or trimodal (See \cref{table:1} for center values), EPD can accurately estimate true parameter distributions $p$ and $K_{\delta}$ (\cref{fig:5}(b-c)). The estimation result of remaining parameters, $\beta$, $\kappa$, $K_{\delta}$, and $\delta$, were provided in (\cref{fig:6}).

\begin{table}

\caption{Parameter values for different distribution types within the target cell-limited model using synthetic data}

\centering
\begin{tblr}{
  hline{1,8} = {-}{0.08em},
  hline{2-3,5} = {-}{},
  colspec={X[1,l]X[1, r]X[1,r]X[1,r]X[1,r]X[1,r]X[1,r]},
}
{Distribution\\type} & $\beta$ \quad ($\times 10^{-4}$) & $p$ & $c$  \quad ($\times10^1$) & $\kappa$ & $\delta$ \quad  ($\times 10^6$) & $K_{\delta}$ \quad   ($\times 10^4$) \\
Unimodal                  & 2.40                                          & 1.60  & 1.30                  & 4.00                      & 1.60                                      & 4.50                                             \\
Bimodal                   & 2.88                                          & 1.44  & 1.82                  & 5.20                      & 1.28                                      & 3.15                                             \\
                          & 2.16                                          & 2.08  & 0.91                  & 3.20                      & 1.76                                      & 4.95                                             \\
Trimodal                  & 2.88                                          & 1.12  & 1.56                  & 4.00                      & 1.44                                      & 4.50                                             \\
                          & 1.68                                          & 2.24  & 1.82                  & 5.60                      & 1.60                                      & 7.20                                             \\
                          & 2.16                                          & 2.08  & 0.78                  & 2.40                      & 1.92                                      & 2.25                                             
\end{tblr}
\label{table:1} 
\end{table}

\section{Application of EPD to real-world datasets}
\subsection{Logistic model}
\begin{figure}[tbhp]
  \centering
  \includegraphics[width=1\textwidth]{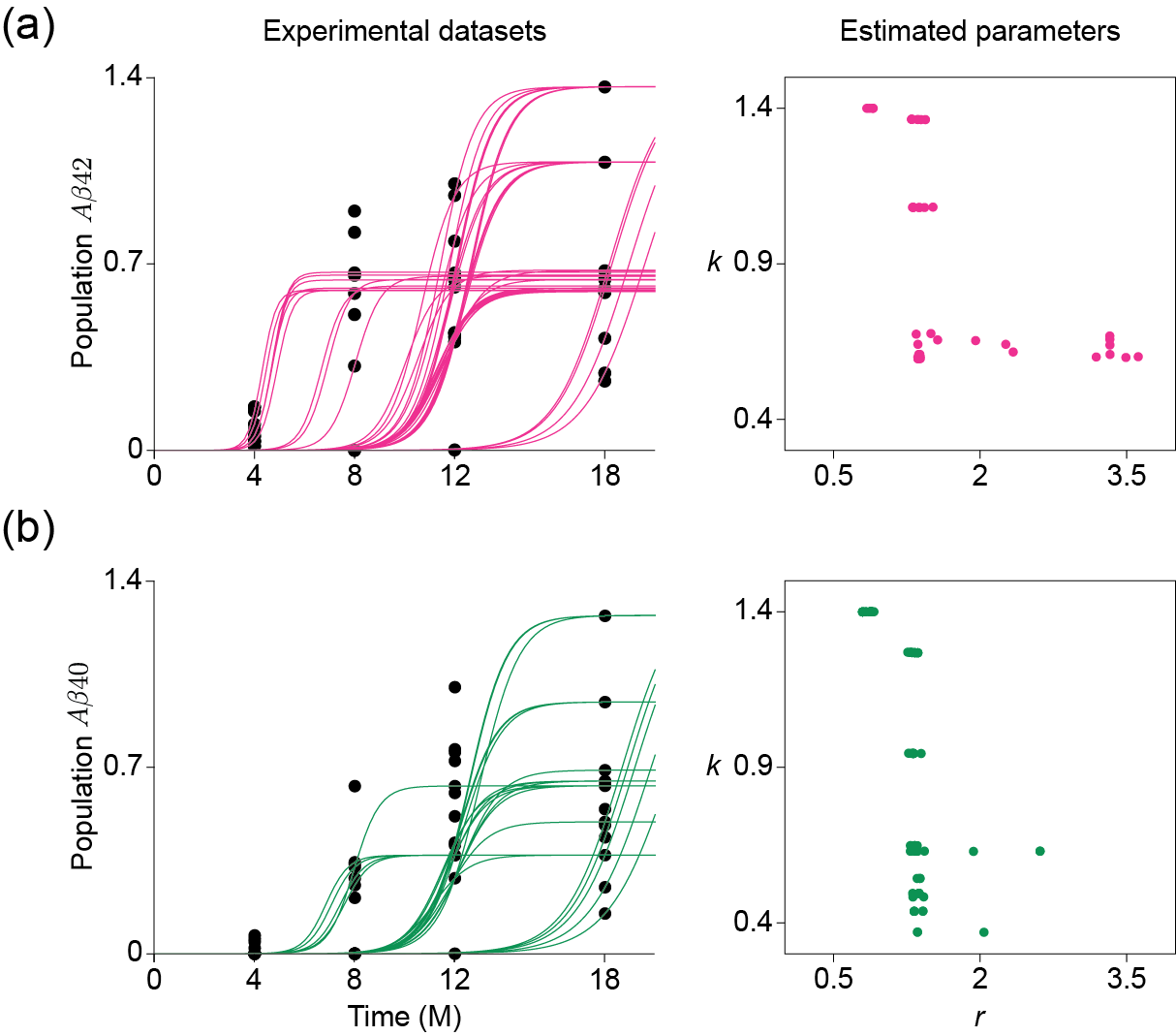}
  \caption{{\bf Estimation results for real experimental datasets on amyloid beta accumulation using a logistic model with EPD.} (a) amyloid beta 40, (b) amyloid beta 42. The left plot shows the accumulation of amyloid beta at 4, 8, 12, and 18 months. We present the estimation results for this dataset using EPD on the right. The left plot also includes some trajectories corresponding to the parameters estimated on the right.}
  \label{fig:7}
\end{figure}

We fitted the logistic model to amyloid-$\beta$ 40 (A$\beta$40) and amyloid-$\beta$ 42 (A$\beta$42) datasets, utilizing them as biomarkers for diagnosing dementia \cite{yada2023few, whittington2018spatiotemporal, hao2022optimal}. In the experimental datasets, the number of  (A$\beta$40) and (A$\beta$42) were recorded at four different time points at 4, 8, 12, and 18 months and each time point had 12-13 independent observation samples (\cref{fig:7}(a-b), left). We then normalized the levels of (A$\beta$40) and (A$\beta$42) (measured in picograms per milliliter), so that the peak value observed in 12-month-old mice was set to 1.0. As the data is RCS type, we utilized EPD for inferring the shape of parameter distribution.  Our results indicate significant heterogeneity in the growth dynamics of amyloid beta, as demonstrated by distinct centers of parameters for both growth rates and population capacities (\cref{fig:7}(a-b), right). As the heterogeneity that shows the progression of amyloid beta accumulation can vary significantly across different population subsets, the estimated parameter distribution implies the importance of personalized diagnostic and therapeutic strategies in combating dementia. Furthermore, we observed that no single parameter could effectively account for the trajectories that included data points at the high rate of amyloid A$\beta$42 at 8 months or A$\beta$40 at 12 months. It implies that when the mouse reaches a certain rate of amyloid A$\beta$40 or 42 early in their life, it cannot survive for a long time. 

\subsection{Target cell-limited model}
\begin{figure}[tbhp]
  \centering
  \includegraphics[width=1\textwidth]{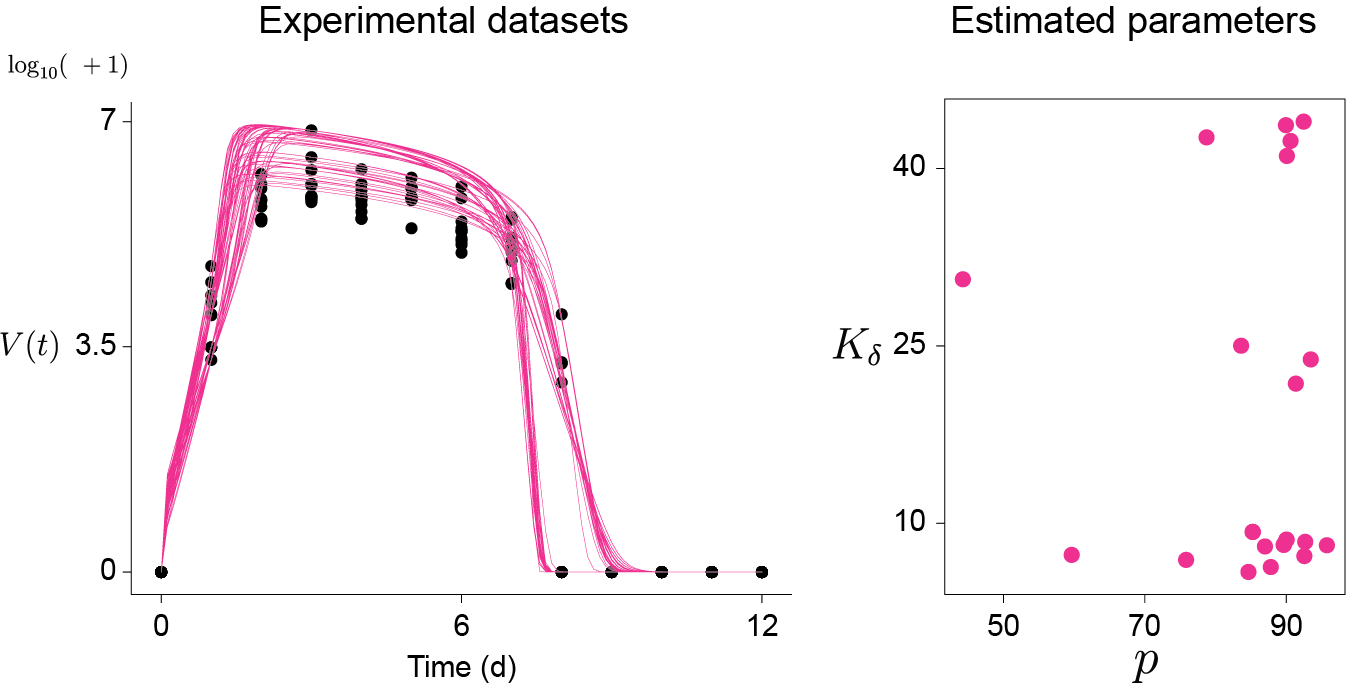}
  \caption{{\bf Estimation results of EPD for a real experimental virus infection dataset using a target cell-limited model with delayed production.} Black dots represent the virus population $V$ over time $t$, the only measurable factor among the four components in the model (left). Artificial trajectories using EPD were simultaneously provided on the same graph (left, red curves). Two parameters corresponding to the artificial trajectories, $p$ and $K_{\delta}$, were estimated (right, red dots). The other parameters, $\beta$, $\kappa$, $K_{\delta}$,and $\delta$, are detailed in \cref{fig:9}}
  \label{fig:8}
\end{figure}

\begin{figure}[tbhp]
  \centering
  \includegraphics[width=\textwidth]{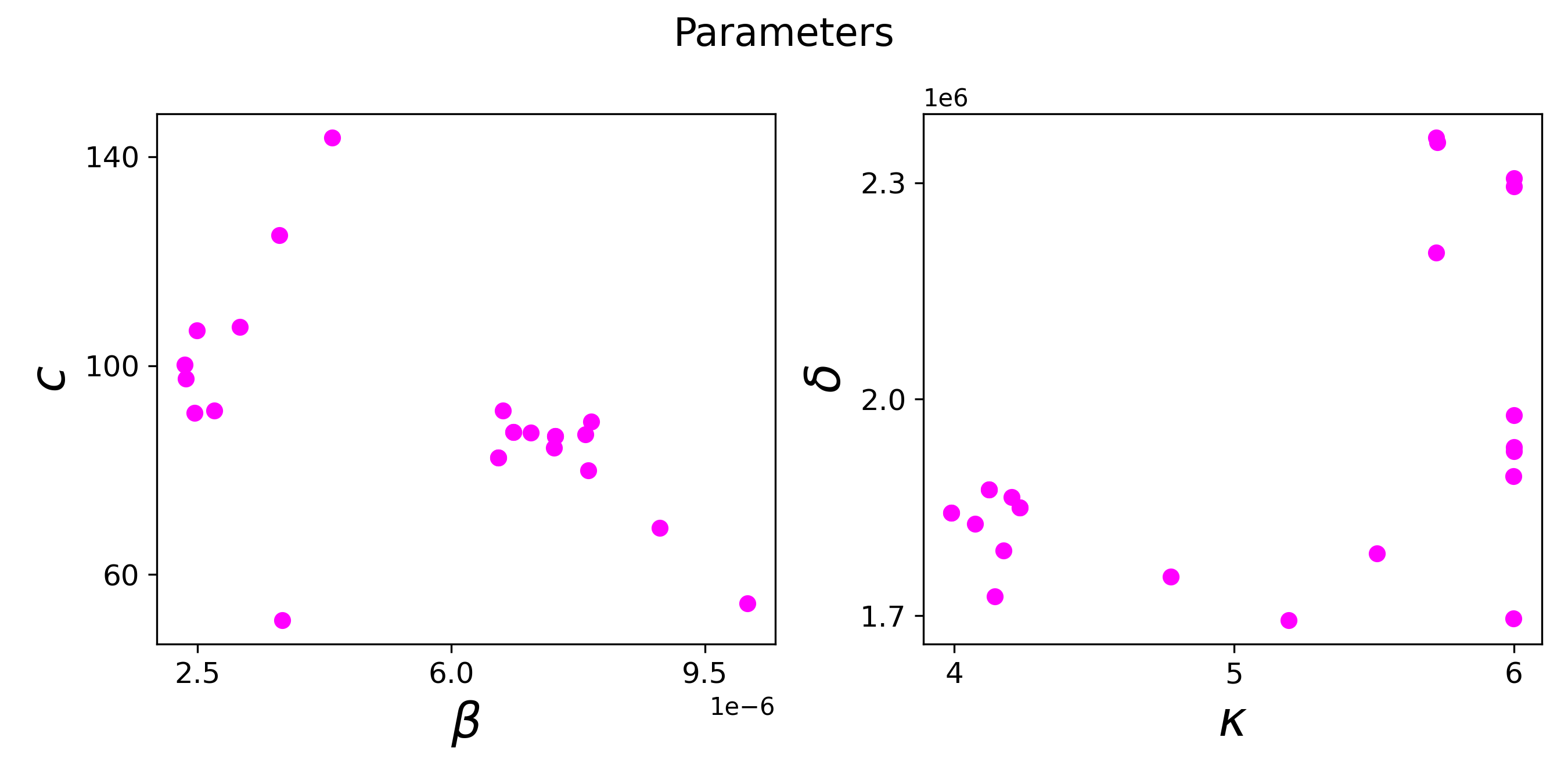}
  \caption{{\bf Estimation results for a real experimental virus infection dataset.} We estimated the four parameters $\beta, c, \kappa, \delta$ that fit the target cell-limited model to the given RCS data \cite{smith2018influenza}. We discovered that the estimated parameter distributions contain heterogeneity for all parameters, similar to the estimates for $p$ and $K_{\delta}$. Unlike previous results in \cite{chung2019parameter}, our findings do not follow the normal distribution shape. Nevertheless, all these predictions could be a reasonable guess because they can reconstruct the trajectories through the equation \eqref{eq:reconstruct_trajectory}.}
  \label{fig:9} 
\end{figure}

We fitted the target cell-limited model to the virus dataset, obtained from \cite{smith2018influenza}. In the dataset, daily viral loads $(V)$ were measured from groups of BALB/cJ mice infected with influenza A/Puerto Rico/8/34 (H1N1) virus (PR8) (\cref{fig:8}, left). The mice received an intranasal administration of a dose of 75 TCID50 of the PR8 virus at the initial time point (t = 0), where TCID50 is the concentration required to infect 50\% of the cell cultures \cite{smith2011effect, smith2013kinetics}. Unlike the previous estimation tasks, only the value of V is observable out of all the populations in the model, thus the parameter estimation was performed using only the viral loads. Data was collected over 12 days, with 10 animals sampled per time point. For faster computation, we only utilized four time points observed at 1, 3, 7, and 8. With this RCS data, we apply EPD for estimating parameter distributions with a target cell-limited model with delayed virus production (\cref{fig:8}, right). Surprisingly, this result shows that two distributions of the parameters $\beta$ and $K_{\delta}$ have at least three centers. The estimation result of remaining parameters, $\beta$, $\kappa$, $K_{\delta}$, and $\delta$, were provided in (\cref{fig:9}). That is, given that EPD not only can accurately predict parameters but also corresponding results show the heterogeneity of parameters. Thus, when it is near the value from the previous research, our results suggest the existence of multiple parameter sets that can represent this dataset, beyond those previously identified parameters. 

\section{Discussion}\label{sec5:Discussion}

\begin{figure}[tbhp]
  \centering
  \includegraphics[width=1\textwidth]{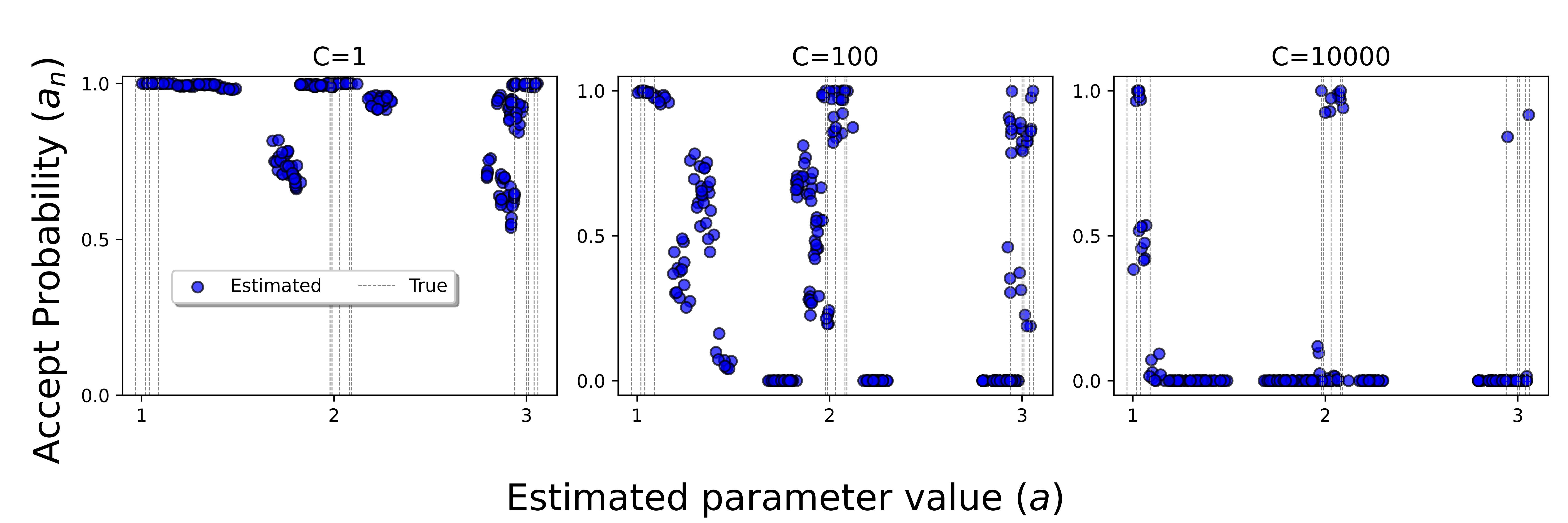}
  \caption{{\bf Accept probability $a_n$ for each scaling factor $C$ in EPD} Three images represent the accept probability of each estimated parameter by AP method within an exponential growth model across three scaling factors: $C=1$ (left), $C=100$ (Middle), $C=10000$ (Right). In all images, estimated parameters from each artificial trajectory are marked by blue dots. The grey lines represent the true value of the growth rate parameter $a$ in the model.}
  \label{fig:10}
  \vspace{-3cm}
\end{figure}
In conclusion, this paper introduces the Estimation of Parameter Distribution (EPD) method for inferring parameter distributions from Repeated Cross-Sectional data in systems modeling. Unlike previous approaches, which often overlooked data heterogeneity and resulted in information loss, EPD enables a more precise and accurate determination of parameter distributions across a variety of systems. By estimating parameter distributions, EPD facilitates a deeper understanding of the underlying dynamics of these systems. Consequently, this paper not only advances our capacity to model and predict system behaviors more effectively but also highlights the critical need to account for data variability and distribution when analyzing complex systems.

We have several limitations for future directions. Since EPD utilized ODESolver, we need to choose a suitable ODESolver that can solve the given dynamical system. Second, parameters were selected from a large set of synthetic trajectories. Through this process, computational costs are proportional to the number of trajectories. Thus, a suitable choice of the number of trajectories is needed.

Determining the appropriate scaling factor $C > 0$, in the accept probability \eqref{eq:accept_probability}, is critical for EPD, because it influences the likelihood of accepting a given parameter $\mathbf{p}$, even when the objective function value $L(\mathbf{p})$ in \eqref{eq:objective} remains unchanged. For example, a higher positive scaling factor leads to only the parameter resulting in a lower objective function value being accepted. While a large scaling factor might suggest that EPD estimations become more precise due to focusing on lower loss values, we should be careful of its magnitude, especially when dealing with parameter heterogeneity (\cref{fig:10}). A large scaling factor will result in the acceptance of only those parameters with minimal loss values. It potentially excludes suitable estimations near certain other centers if there exist noticeable differences in loss values between parameters around different centers.

Finally, we can consider other transformations when constructing accept probability. We currently apply the logistic transformation to the loss function. Without any transformation, the acceptance probability is directly proportional to the loss function value, which has no significant distinction between parameters with different loss values. The logistic transformation helped EPD to select a parameter with much less objective function value. In future research, we will clarify whether this transformation is optimal by providing rigorous proof or conducting various experiments with other transformations.

\bibliographystyle{siamplain}
\newpage
\bibliography{references}

\end{document}